\documentclass{article}
\usepackage{spconf,amsmath,graphicx}

\usepackage{newtxtext,newtxmath}

\usepackage{tikz}
\usetikzlibrary{arrows.meta,positioning,fit,backgrounds,calc}
\usepackage{booktabs}
\usepackage{multirow}
\usepackage{siunitx}
\usepackage[hidelinks]{hyperref}
\usepackage[capitalise,noabbrev]{cleveref}

\usepackage{xcolor}
\newcommand{\todo}[1]{\textcolor{red}{[\textbf{TODO:} #1]}}

\title{\todo{TITLE, ALL CAPS}}

\name{\todo{Author Names}}
\address{\todo{Affiliation}}

\crefname{figure}{Fig.}{Figs.}
\Crefname{figure}{Fig.}{Figs.}
\usepackage{tipa}
\newcommand{\ipa}[1]{\textipa{#1}}
\title{PVSync: A Unified Lip-Sync Expert for Timing and Articulation}
\name{Kevin Stephen$^{1,2,\star,\dagger}$ \qquad Varun Menon$^{2,\star}$ \qquad Timo Mertens$^{2}$ \qquad Nikita Drobyshev$^{2}$}
\address{$^{1}$University of Southern California \qquad $^{2}$Cantina Inc.\\[2pt]
{\small $^{\star}$Equal contribution. \quad $^{\dagger}$Work done during an internship at Cantina Inc.}}
\makeatletter
\long\def\@makecaption#1#2{\vskip 7pt\setbox\@tempboxa\hbox{#1. #2}%
  \ifdim \wd\@tempboxa >\hsize #1. #2\par \else \hbox to\hsize{\hfil\box\@tempboxa\hfil}\fi
  \vskip 3pt}
\makeatother
\makeatletter
\renewcommand\section{\@startsection{section}{1}{\z@}{-2.0ex plus -0.5ex minus -.2ex}{1.2ex plus .2ex}{\normalfont\normalsize}}
\renewcommand\subsection{\@startsection{subsection}{2}{\z@}{-1.6ex plus -0.4ex minus -.2ex}{0.8ex plus .2ex}{\normalfont\normalsize}}
\makeatother

\begin{document}
\ninept   %
\makeatletter
\g@addto@macro\small{\abovedisplayskip=3pt plus 1pt minus 1pt
  \belowdisplayskip=3pt plus 1pt minus 1pt
  \abovedisplayshortskip=1pt \belowdisplayshortskip=2pt}
\makeatother
\let\normalsize\small  %
\normalsize
\let\oldthebibliography\thebibliography
\def\thebibliography#1{\oldthebibliography{#1}\setlength{\itemsep}{1pt plus 0.5pt}\setlength{\parsep}{0pt}}
\raggedbottom
\maketitle

\begin{abstract}

Lip movements can match the timing of speech without matching the spoken
sounds. We introduce PVSync, a unified model for audio--visual
offset estimation and phoneme-level articulation scoring. PVSync combines
window-level contrastive learning for synchronisation with a phoneme-level
articulation objective that aligns audio and video embeddings of the same
viseme class across clips. Visemes group phonemes with similar visible
articulation. Viseme labels are derived automatically from forced-aligned
transcripts, without manual annotations.
On videos from \num{13} talking-head video generation models, PVSync
matches human rankings of lip-sync quality more closely than LSE-C,
achieving a Spearman correlation of \num{0.83} versus \num{0.34}.
On an automatically constructed benchmark from held-out speech, PVSync
distinguishes viseme-matched from mismatched audio--visual pairs with
an ROC AUC of \num{0.91}. PVSync also recovers the offset within one frame on \SI{92.7}{\percent} of
held-out in-the-wild clips, against \SI{87.4}{\percent} for MTD-VocaLiST and
\SI{80.0}{\percent} for SyncNet.
Code and benchmark data will be released upon acceptance.

\end{abstract}


\section{Introduction and Related Work}
\label{sec:intro}

Accurate lip-sync requires more than temporal alignment. A generated mouth
can move in time with speech yet fail to close for /p/ 
or form the lip--tooth contact required for /f/. Clip-level scores may
obscure these brief errors, motivating evaluation of both timing and
local articulation.

\begin{figure}[!t]
\centering
\includegraphics[width=\columnwidth]{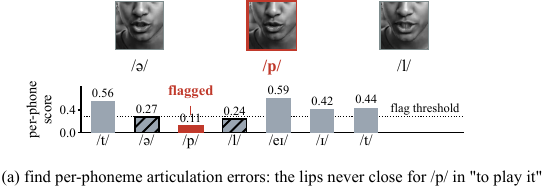}\\[5pt]
\includegraphics[width=\columnwidth]{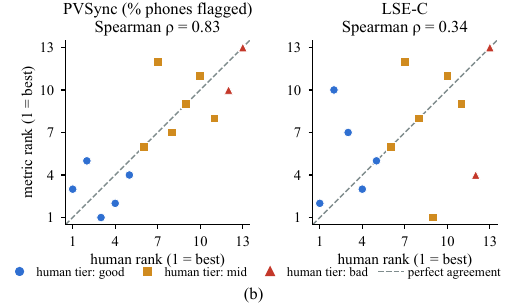}
\caption{\textbf{(a)}~A local articulation error caught on a generated clip: the per-phone
score of the /p/, where the lips never close, falls below the flag threshold. \textbf{(b)}~Human
rank against metric rank for thirteen talking-head video generation models.}
\label{fig:space}
\vspace{-7pt}
\end{figure}

LSE-C, derived from SyncNet~\cite{chung2016out}, is widely used to evaluate
talking-head video generation models including Wav2Lip~\cite{prajwal2020wav2lip}
and LatentSync~\cite{li2024latentsync}. Yet high scores do not establish
correct phoneme-level articulation.

We introduce \textbf{PVSync}, a unified model for temporal offset
estimation and phoneme-level articulation scoring in a shared audio--visual
embedding. A contrastive objective aligns same-viseme audio
and video spans across clips, alongside window-level synchronisation
learning. Labels come from forced-aligned transcripts, without manual annotation. A final projection-head fine-tuning stage
improves temporal precision while retaining articulation-sensitive
representations.

On offset-corrected videos from \num{13} talking-head video generation models, PVSync
agrees with the human ranking of the models at Spearman $\rho = \num{0.83}$,
versus \num{0.34} for LSE-C (\num{1620} ratings by \num{180} blind raters,
\cref{fig:space}b). PVSync
achieves ROC AUC \num{0.91} on an automatically constructed benchmark
of viseme-matched and mismatched audio--visual pairs from held-out speech
(\cref{tab:pair}). With \num{12}-frame context, it recovers offsets
within one frame at \SI{92.7}{\percent} accuracy, versus
\SI{87.4}{\percent} for MTD-VocaLiST and \SI{80.0}{\percent} for SyncNet
(\cref{tab:kcontext}).

VocaLiST~\cite{kadandale2022vocalist},
MTD-VocaLiST~\cite{mtdvocalist2024},
Synchformer~\cite{iashin2024synchformer} and
SyncLipMAE~\cite{ling2025synclipmae} focus on offset estimation.
Jawade et al.~\cite{netflix2025ranking} address dubbed-video content
mismatch through ranking-based training, but produce a clip-level score.
PhoVis~\cite{gupta2024phovis} and FluentLip's phoneme error rate~\cite{liu2025fluentlip}
score phoneme-level errors with landmark templates and an AV-HuBERT lip-reader,
but both compare the mouth with transcribed phoneme labels, not the audio itself,
and neither estimates an offset. PVSync unifies learned cross-modal phoneme-level scoring
and offset estimation.

Our contributions are:
\begin{figure*}[!t]
\centering
\includegraphics[width=0.43\textwidth]{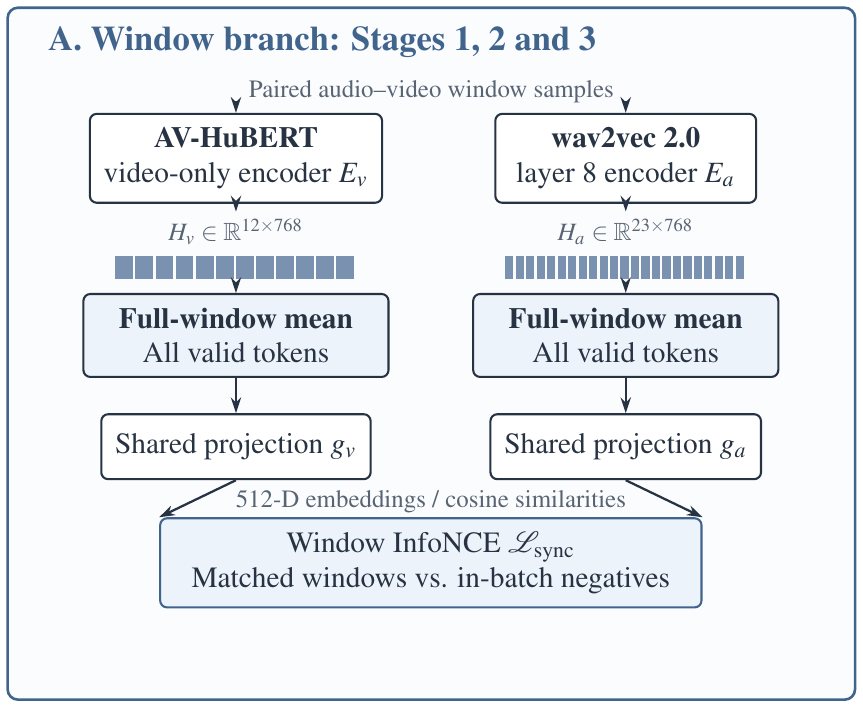}\hfill
\includegraphics[width=0.43\textwidth]{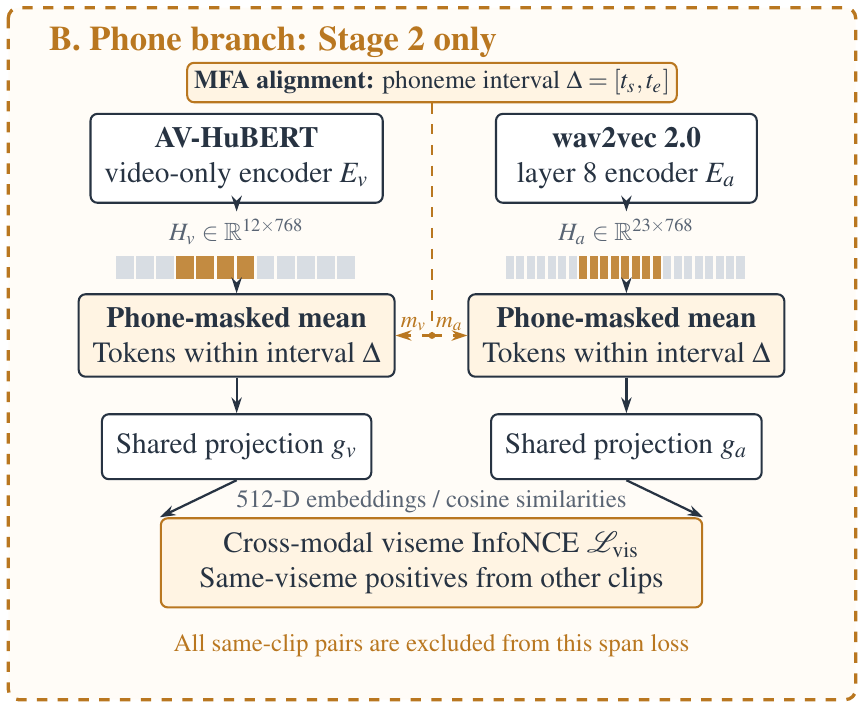}
\caption{Training paths; encoders and heads are shared. \textbf{A.}~Window
branch, every stage: a \num{12}-frame window's tokens are mean-pooled,
projected and scored by cosine under $\mathcal{L}_{\text{sync}}$; stage~3
trains only the heads. \textbf{B.}~Phone branch, stage~2 only: a
forced-aligned interval~$\Delta$ selects the same tokens in both towers
before the mean, and $\mathcal{L}_{\text{vis}}$ pulls the span toward
same-viseme spans of other clips.}
\label{fig:stages}
\end{figure*}

\begin{itemize}
\setlength{\itemsep}{0pt}

\item \textbf{A unified model for synchronisation and articulation
assessment.} One audio--visual embedding supports offset estimation
and phoneme-level lip-sync scoring (\cref{fig:stages}).

\item \textbf{A phoneme--viseme-aware contrastive objective.}
Same-viseme spans from different clips provide supervision for
learning articulation-sensitive representations without manual annotation (\cref{ssec:vis}).

\item \textbf{A benchmark and human evaluation for articulation-aware
lip-sync scoring.} Automatically constructed viseme-matched and
mismatched pairs assess articulation scoring; a study of \num{13}
talking-head generation models demonstrates improved agreement with human lip-sync rankings.

\end{itemize}

\section{Method}
\label{sec:method}

\subsection{Architecture}
\label{ssec:arch}
A clip is scored in \num{12}-frame (\SI{480}{\ms}) windows at 25~fps.
The video tower is AV-HuBERT Base \cite{shi2022avhubert} video-only, reading
$88\times88$ grayscale mouth ROIs; the audio tower is wav2vec2-base
\cite{baevski2020wav2vec2} read at layer~8 over the matching \num{7680}
samples. The video tower emits one \num{768}-d token per frame and the audio tower one
every \SI{20}{\ms}, so a window gives \num{12} video tokens and \num{23} audio
tokens, $\mathbf{H}^v\in\mathbb{R}^{12\times768}$ and
$\mathbf{H}^a\in\mathbb{R}^{23\times768}$ (\cref{fig:stages}). The audio count
is \num{23} rather than \num{24} because the front end has a \num{400}-sample
receptive field at a \num{320}-sample stride and no padding, so
$\lfloor(7680-400)/320\rfloor+1$ tokens fit and the last partial stride is
dropped. In each
tower, the tokens of the window are mean-pooled over time and the pooled
\num{768}-d vector is projected by a two-layer head into a shared
\num{512}-dimensional space. The similarity score is the cosine similarity
$S=\cos(\mathbf{f}^v,\mathbf{f}^a)$ of the two projections.

\subsection{Training}
\subsubsection{Stage 1: audio-visual contrastive pre-training}
Stage~1 trains both towers with InfoNCE at temperature $\tau$ on batches of
$B$ \num{12}-frame windows sampled at random from the training set. For video
window $i$ the positive is its own audio and the negatives are the audio
windows of the other $B{-}1$ clips in the batch, so $j$ in the denominator
runs over all $B$ windows.
\begin{equation}
\mathcal{L}_{\text{sync}} = -\log
\frac{\exp(\cos(\mathbf{f}^v_i,\mathbf{f}^a_i)/\tau)}
     {\sum_j \exp(\cos(\mathbf{f}^v_i,\mathbf{f}^a_j)/\tau)} .
\end{equation}
The stage-1 model recovers the offset more
accurately than every baseline in \cref{tab:kcontext}, before any
viseme supervision.

\subsubsection{Stage 2: cross-modal viseme alignment}
\label{ssec:vis}
Stage~2 adds a second contrastive term defined on phones rather than windows.
We force-align each training transcript to its audio \cite{mcauliffe2017mfa},
which gives every phone a start and end time, and map each phone to one of
eight viseme classes (the rows of \cref{tab:pair_confusion}). Each phone's interval selects the tokens it covers in
both towers, a token counting as covered when the centre of the time it
represents lies inside the interval. Those tokens are mean-pooled within their
tower and projected through the tower's head, exactly as a whole window is in
\cref{ssec:arch}, giving one embedding per modality for the phone, its \emph{video span} and its
\emph{audio span}; together we call them a \emph{phone span}. We use phones of at least two frames (\SI{80}{\ms}),
so a span covers at least two video tokens and four audio tokens (the phone
branch of \cref{fig:stages}). For a video span $i$, the positives are the audio spans of the same viseme class taken from
other clips,
\begin{equation}
P(i)=\{\,j : v_j = v_i \;\wedge\; c_j \neq c_i \,\},
\end{equation}
where $v$ is the viseme class and $c$ the clip. All other spans from other clips
are negatives, and spans from the same clip are excluded from the denominator
as well, so no term in the loss can be satisfied by anything a true pair shares
apart from the viseme, such as the speaker's voice or the recording. With a video span as the anchor and audio spans as candidates, the
video-to-audio term is
\begin{equation}
\mathcal{L}_{\text{vis}}^{v\to a}(i) = \frac{-1}{|P(i)|}\sum_{p\in P(i)}
\log \frac{\exp(\cos(\mathbf{f}^v_i,\mathbf{f}^a_p)/\tau_v)}
          {\sum_{j:\,c_j\neq c_i}\exp(\cos(\mathbf{f}^v_i,\mathbf{f}^a_j)/\tau_v)}.
\label{eq:vis_dir}
\end{equation}

The viseme loss averages \cref{eq:vis_dir} and its symmetric audio-to-video
counterpart $\mathcal{L}_{\text{vis}}^{a\to v}(i)$ over the $B_s$ spans of a batch
and over the two directions,
\begin{equation}
\mathcal{L}_{\text{vis}} = \tfrac{1}{2}\Big(
\tfrac{1}{B_s}\sum_i \mathcal{L}_{\text{vis}}^{v\to a}(i) +
\tfrac{1}{B_s}\sum_i \mathcal{L}_{\text{vis}}^{a\to v}(i)\Big),
\end{equation}
and the stage-2 objective is $\mathcal{L}_{\text{sync}} + \lambda\mathcal{L}_{\text{vis}}$,
with $\mathcal{L}_{\text{sync}}$ computed as in stage~1 on its own window batch.
$\mathcal{L}_{\text{sync}}$ is kept in stage~2 because $\mathcal{L}_{\text{vis}}$
alone destroys temporal alignment (\cref{tab:kcontext}).

\subsubsection{Stage 3: synchronisation fine-tuning}
The ablation in \cref{tab:kcontext} shows that $\mathcal{L}_{\text{vis}}$ alone
destroys timing (\num{0.210} at $K=12$), and even with $\mathcal{L}_{\text{sync}}$
alongside it stage~2 gives up a little (\num{0.906} to \num{0.892}). Stage~3
therefore freezes both encoders, which maintains the viseme clustering learnt in
stage~2 (\cref{tab:pair}), and fine-tunes only the two projection heads under
$\mathcal{L}_{\text{sync}}$, which brings timing to \num{0.927}, above stage~1.
The same fine-tuning applied to encoders trained with $\mathcal{L}_{\text{vis}}$
alone reaches only \num{0.827}, below stage~1 and below MTD-VocaLiST.

\section{Experiments}
\label{sec:setup}
\label{sec:results}

\subsection{Implementation details}
\label{ssec:impl}

\noindent\textbf{Data.} We train on VoxCeleb2 \cite{chung2018voxceleb2} (\num{967}k
clips, \num{2073} hours), an internal set of talking-head recordings, and six
public datasets: Hallo3 \cite{cui2024hallo3}, CelebV-HQ \cite{zhu2022celebvhq},
CelebV-Text \cite{yu2023celebvtext}, VFHQ \cite{xie2022vfhq}, MEAD
\cite{wang2020mead} and RAVDESS \cite{livingstone2018ravdess}. The internal
and public sets total \num{420}k clips (\num{1055} hours) after a SyncNet
filter that drops clips with SyncNet confidence below \num{4} or an offset more than
three frames from the dataset mode, and shifts the rest to their detected
offset, so the audio of every kept clip is aligned to its video. VoxCeleb2 is
not filtered. Stages~2 and~3 fine-tune on the filtered internal and
public sets only, and stage~2 also draws \num{1.88}M forced-aligned phone
spans \cite{mcauliffe2017mfa} from \num{77}k of the filtered clips. The in-the-wild test
clips of \cref{ssec:offset} were split off before training; HDTF, LRS2 and the
generated clips appear in no training set.

\noindent\textbf{Training.} All stages train on eight H100 GPUs with an effective
batch of $B=512$, a learning rate of \num{1e-5} for the projection heads and
\num{5e-6} for unfrozen encoder blocks, and audio augmented with pitch shift,
gain and additive noise. \emph{Stage~1} freezes the convolutional front end of each encoder and trains all
twelve AV-HuBERT transformer blocks and wav2vec2 blocks 0--7 (\num{143}M trainable
parameters) for \num{300}k steps at $\tau=0.1$.
\emph{Stage~2} continues for \num{20}k steps and adds a batch of $B_s=64$ phone
spans per GPU, eight per viseme class, so every anchor has up to seven positives
and about \num{56} negatives after same-clip masking; $\lambda=1$, $\tau_v=0.1$.
Each step runs the encoders once over the window batch and once over the span
batch and takes one gradient step on $\mathcal{L}_{\text{sync}}+\lambda\mathcal{L}_{\text{vis}}$.
\emph{Stage~3} freezes both encoders and trains only the two projection heads
(\num{1.31}M parameters, \SI{0.7}{\percent} of the model) for \num{10}k steps
under $\mathcal{L}_{\text{sync}}$ at $\tau=0.2$.

\subsection{Offset recovery}
\label{ssec:offset}
\begin{table}[t]
\centering
\caption{Lip-synchronisation accuracy (offset correct within one frame) on \num{3698}
held-out clips at context window $K$ (\cref{ssec:offset}). Best in bold, second best underlined.}
\label{tab:kcontext}
\small  %
\setlength{\tabcolsep}{2pt}
\begin{tabular}{lcccc}
\toprule
Model (input frames) & $K{=}12$ & $K{=}16$ & $K{=}20$ & $K{=}25$ \\
\midrule
PVSync stage 1 (12)              & \underline{0.906} & \underline{0.944} & \underline{0.962} & 0.975 \\
PVSync stage 2 (12)              & 0.892 & 0.938 & 0.960 & \underline{0.976} \\
\textbf{PVSync stage 3 (12)}     & \textbf{0.927} & \textbf{0.951} & \textbf{0.968} & \textbf{0.981} \\
\addlinespace[1pt]
\addlinespace[2pt]
\quad $\mathcal{L}_{\text{vis}}$ alone (12)          & 0.210 & 0.256 & 0.322 & 0.396 \\
\quad + head fine-tuning under $\mathcal{L}_{\text{sync}}$ (12) & 0.827 & 0.902 & 0.939 & 0.964 \\
\addlinespace[2pt]
SyncNet \cite{chung2016out} (5)          & 0.800 & 0.883 & 0.929 & 0.960 \\
MTD-VocaLiST \cite{mtdvocalist2024} (5)  & 0.874 & 0.925 & 0.954 & 0.973 \\
StableSyncNet \cite{li2024latentsync} (16) & ---   & 0.906 & 0.936 & 0.958 \\
\bottomrule
\end{tabular}
\end{table}

\begin{table}[t]
\centering
\caption{Per-window offset error in frames at \SI{25}{fps}. Each of the \num{528757}
windows predicts an offset from its own input alone (\cref{ssec:offset}). Mean and std are over all windows,
std$_{16}$ the std when the \num{16}-frame context of \cref{tab:kcontext} is used.
Best in bold, second best underlined.}
\label{tab:errspread}
\small  %
\setlength{\tabcolsep}{2.6pt}
\begin{tabular}{lrrrrrr}
\toprule
Model & mean & std & std$_{16}$ & \multicolumn{3}{c}{\% of windows with $|$error$|$} \\
\cmidrule(lr){5-7}
 & & & & ${=}0$ & ${\le}1$ & ${\ge}5$ \\
\midrule
PVSync stage 1          & $-0.14$ & \textbf{1.72} & \textbf{1.49} & 52.3 & 90.3 & \textbf{2.2} \\
PVSync stage 2          & $+0.04$ & 2.35 & 1.93 & \underline{53.9} & 89.2 & 4.6 \\
\textbf{PVSync stage 3} & $+0.03$ & \underline{1.93} & \underline{1.65} & \textbf{63.7} & \textbf{92.6} & \underline{3.1} \\
\addlinespace[1pt]
SyncNet                 & $-0.05$ & 6.35 & 3.06 & 27.2 & 51.6 & 35.4 \\
MTD-VocaLiST            & $-0.40$ & 5.06 & 2.14 & 32.8 & 66.8 & 22.5 \\
StableSyncNet           & $+0.47$ & 2.29 & 2.29 & 44.8 & \underline{91.2} & 4.2 \\
\bottomrule
\end{tabular}
\end{table}

\begin{table}[t]
\centering
\caption{Lip-synchronisation accuracy on the LRS2 test set (\num{1243} clips, correct within one
frame, VocaLiST's protocol and code) at context window $K$. Published rows are from
VocaLiST's Table~1 \cite{kadandale2022vocalist} and are trained on LRS2; the lower rows are
our runs. Best in bold, second best underlined.}
\label{tab:lrs2}
\small  %
\setlength{\tabcolsep}{5pt}
\begin{tabular}{lccc}
\toprule
Model & $K{=}13$ & $K{=}15$ & $K{=}16$ \\
\midrule
SyncNet \cite{chung2016out}                & 94.5 & 96.1 & --- \\
Perfect Match                              & 98.7 & 99.1 & --- \\
AVST                                       & \underline{99.3} & \underline{99.6} & --- \\
VocaLiST                                   & \textbf{99.6} & \textbf{99.8} & --- \\
\addlinespace[1pt]
SyncNet, our run                           & 95.0 & 96.7 & 97.3 \\
MTD-VocaLiST \cite{mtdvocalist2024}, our run & 99.2 & 99.5 & \textbf{99.6} \\
StableSyncNet \cite{li2024latentsync}, our run & --- & --- & \underline{98.0} \\
\textbf{PVSync stage 3}, not trained on LRS2         & 99.2 & 99.5 & \textbf{99.6} \\
\bottomrule
\end{tabular}
\end{table}

\noindent\textbf{Protocol.} We follow VocaLiST's evaluation protocol
\cite{kadandale2022vocalist}. Each test clip's audio is aligned to its
SyncNet-detected offset (\cref{ssec:impl}), so the true offset is zero. Each model
judges a fixed-length input, \num{12} frames for PVSync. To decide over a longer
context of $K$ frames, we shift the audio by a candidate offset, score every
input-length window inside those $K$ frames one frame apart, and average the
scores. Over the candidate offsets from $-15$ to $+15$ frames, the predicted
offset is the one with the highest average, and it is correct within one frame of
zero. The SyncNet row is therefore a self-consistency baseline.
The test set is \num{3698} held-out clips from Hallo3, CelebV-Text, CelebV-HQ,
VFHQ and RAVDESS. Baselines are SyncNet \cite{chung2016out}, MTD-VocaLiST
\cite{mtdvocalist2024} (standing in for VocaLiST, whose weights are unavailable) and StableSyncNet \cite{li2024latentsync}, all run through
our pipeline.

\noindent\textbf{Results.} At a \num{12}-frame context window PVSync stage~3
reaches a synchronisation accuracy of \SI{92.7}{\percent}, against
\SI{87.4}{\percent} for MTD-VocaLiST and \SI{80.0}{\percent} for SyncNet
(\cref{tab:kcontext}). It is also more consistent than every baseline, with a
per-window error standard deviation of \num{1.93} frames against \num{6.35} for
SyncNet (\cref{tab:errspread}). SyncNet trails on a test set aligned by its own offsets because its
estimate is reliable once pooled over a clip but not window by window. On the LRS2 test set \cite{afouras2018lrs2},
which no PVSync stage trained on, PVSync lands within \SI{0.4}{pp} of the
in-domain state of the art (\cref{tab:lrs2}). AVST, VocaLiST, MTD-VocaLiST and PVSync all exceed \SI{99}{\percent} at $K{=}15$, so LRS2 no longer separates them, which is why we report on in-the-wild clips.

\subsection{Reading visemes}
\label{ssec:loc}

\noindent\textbf{Per-phone score.} A \num{12}-frame (\SI{480}{\ms}) window slides over
the clip with a stride of 1 frame, taking the same \SI{480}{\ms} of video and
audio at every position. Within each window, forced alignment (\cref{ssec:vis})
gives each phone a video span and an audio span, the tokens its interval covers
in each tower. For every phone whose interval is at least \SI{80}{\percent}
covered by the window, the two spans are embedded and their cosine similarity is
recorded. A phone's score is the mean of these cosines across all windows it
appears in, and it is flagged when the score falls below a threshold.

\noindent\textbf{Benchmark.} The per-phone score compares a phone's video span with
its own audio span. To test that it responds to phoneme--viseme agreement, we pair
video spans with the audio spans of other phones and ask whether the score
separates matching visemes from mismatched ones. We build the benchmark on HDTF
\cite{zhang2021hdtf}, \num{805} clips of \num{292} speakers, on which no stage of
PVSync and none of the baselines was trained. The footage is clean and frontal,
the speech well articulated, and each clip's audio is aligned to its
SyncNet-detected offset, so every video span shows the sound of its phone being
produced. Phones come from Whisper transcripts force-aligned with MFA and are
mapped to the eight viseme classes of \cref{tab:pair_confusion}. The bank holds
\num{2163} phone spans longer than \SI{80}{\ms}, \numrange{240}{330} per class and
at most seven per clip. Every video span is scored against every audio span from
a different speaker, \num{4.6}M pairs in all, so the two spans of a pair share
nothing but, at most, their viseme class. A pair whose spans share a class is
\emph{good} and should score above the flag threshold. Any other pair is
\emph{bad} and should fall below it. Because the labels come from forced
alignment, any transcribed speech dataset yields such a benchmark at any size
with no human annotation.

\begin{table}[t]
\centering
\caption{Viseme-mismatch flagging on HDTF: each of \num{2163} video spans scored against every
audio span from a different speaker (\num{4.6}M pairs; \cref{ssec:loc}). Brackets: \SI{95}{\percent} clip-clustered
bootstrap. Best in bold, second best underlined.}
\label{tab:pair}
\small  %
\setlength{\tabcolsep}{0.8pt}
\begin{tabular}{lcccccc}
\toprule
 & ROC AUC & \multicolumn{3}{c}{bad pairs flagged (\%)} & \multicolumn{2}{c}{retrieval@$k$} \\
\cmidrule(lr){3-5}\cmidrule(lr){6-7}
 & & @10 & @5 & @1 & $k{=}1$ & $k{=}2$ \\
\midrule
PVSync stage 1                      & 0.694 [.684,.705] & 21.3 & 10.8 & 1.9  & 0.626 & 0.802 \\
$\mathcal{L}_{\text{vis}}$ only     & \textbf{0.964} [.959,.969] & \textbf{90.5} & \textbf{82.9} & \textbf{57.8} & \underline{0.807} & \underline{0.933} \\
PVSync stage 2                      & \underline{0.960} [.955,.965] & \underline{89.7} & \underline{80.8} & \underline{52.8} & \textbf{0.811} & \textbf{0.941} \\
PVSync stage 3                      & 0.909 [.901,.918] & 73.0 & 59.6 & 28.5 & 0.741 & 0.900 \\
\bottomrule
\end{tabular}
\end{table}

\noindent\textbf{Results.} Sweeping the flag threshold gives an ROC curve of bad pairs
caught against good pairs wrongly flagged. \Cref{tab:pair} reports the area under it (ROC AUC, chance
\num{0.5}) and the share of bad pairs flagged at false-flag rates of
\SI{10}{\percent}, \SI{5}{\percent} and \SI{1}{\percent}. Stage~1 reaches an ROC AUC of
\num{0.694} and flags \SI{21}{\percent} of bad pairs at \SI{10}{\percent};
stage~2 reaches \num{0.960} and \SI{90}{\percent}, and stage~3, the shipped
model, \num{0.909} and \SI{73}{\percent}. Training with
$\mathcal{L}_{\text{vis}}$ alone scores highest (\num{0.964}, \SI{91}{\percent})
but loses the offset almost entirely (\cref{tab:kcontext}), so we do not
consider it further.

\begin{table}[t]
  \centering
  \small  %
  \setlength{\tabcolsep}{1pt}
  \caption{Retrieval confusion (\%) at $k{=}1$ (\cref{tab:pair}), stage~2/stage~3 in each
    cell. Rows are the true class; the bold diagonal is each class's retrieval accuracy,
    and cells below \num{0.5}\% are $\cdot$. Classes: bilabial \ipa{/p b m/}; labiodental \ipa{/f v/}; dental/alveolar
    \ipa{/t d n s z l T D S Z tS dZ/}; velar \ipa{/k g N h/}; rounded \ipa{/oU u w U OI O/};
    spread \ipa{/i I eI E j/}; open \ipa{/A \ae\ 2 aU aI/}; rhotic \ipa{/r \textrhookrevepsilon/}.}
  \label{tab:pair_confusion}
  \begin{tabular}{lccccccccr}
    \toprule
    class & bil & lab & d/a & vel & rnd & rho & opn & spr & $n$ \\
    \midrule
    bilabial    & \textbf{98}/\textbf{98} & $\cdot$/$\cdot$ & $\cdot$/$\cdot$ & $\cdot$/$\cdot$ & $\cdot$/$\cdot$ & 1/1 & $\cdot$/$\cdot$ & $\cdot$/$\cdot$ & 252 \\
    labiodental & 2/2 & \textbf{97}/\textbf{97} & 1/1 & $\cdot$/$\cdot$ & $\cdot$/$\cdot$ & $\cdot$/$\cdot$ & $\cdot$/$\cdot$ & $\cdot$/$\cdot$ & 266 \\
    dental/alv. & 3/3 & 1/2 & \textbf{78}/\textbf{78} & 5/3 & 2/6 & 5/2 & 3/3 & 2/2 & 240 \\
    velar       & 3/4 & $\cdot$/$\cdot$ & 6/11 & \textbf{65}/\textbf{33} & 6/13 & 5/4 & 4/12 & 11/21 & 290 \\
    rounded     & 1/1 & $\cdot$/$\cdot$ & $\cdot$/$\cdot$ & 1/1 & \textbf{82}/\textbf{85} & 11/8 & 4/4 & $\cdot$/$\cdot$ & 330 \\
    rhotic      & 1/2 & 1/3 & 3/5 & 2/3 & 15/27 & \textbf{75}/\textbf{56} & 1/3 & 2/1 & 264 \\
    open        & $\cdot$/2 & $\cdot$/1 & 1/3 & 3/3 & 2/2 & 5/4 & \textbf{78}/\textbf{75} & 10/10 & 250 \\
    spread      & $\cdot$/1 & $\cdot$/$\cdot$ & 3/7 & 3/1 & 1/2 & 2/2 & 13/14 & \textbf{77}/\textbf{74} & 271 \\
    \bottomrule
  \end{tabular}
\end{table}

\noindent\textbf{Retrieval.} For each of the eight viseme classes we average the
embeddings of its audio spans into one point, the class centroid, leaving out
spans from the clip being queried. A video span is classified correctly at $k$
when its own viseme class is among the $k$ centroids closest to it by cosine;
retrieval@$k$ is this accuracy over all video spans (\cref{tab:pair}, last two columns; chance
\num{0.125} and \num{0.25}). Stage~3 scores \num{0.74} at $k{=}1$ and \num{0.90}
at $k{=}2$. The audio clusters themselves are well
formed, since a nearest-centroid classifier on the audio embeddings alone reaches an
accuracy of \num{0.94} after stage~3, against \num{0.70} after stage~1. The retrieval
confusion at $k{=}1$ (\cref{tab:pair_confusion}) shows the lip-explicit
classes, labiodental (/f, v/) and bilabial (/p, b, m/), retrieved almost
perfectly, with the errors falling between classes that look alike on the lips. Open and spread
vowels are exchanged, because a vowel such as \ipa{/\ae/} (``cat'') opens the mouth
only slightly more than a spread vowel such as \ipa{/E/} (``bed''). Rhotic
and rounded are exchanged, because /r/ is produced with rounded lips. Velar is
the weakest class and scatters over its neighbours, because /k/ and /g/ have no
lip gesture of their own and the mouth during them shows the neighbouring
vowel. In all, \SI{70}{\percent} of stage-3 errors fall on one of these adjacent
classes.

\subsection{Agreement with human judgments on generated video}
\label{ssec:generated}
\label{ssec:human}
\begin{table}[t]
\centering
\caption{ROC AUC of each chunk score against the human majority label (\num{400} chunks of
\SI{480}{\ms}, generated video) and flagged chunks caught at a \SI{16}{\percent} false-flag
rate. Brackets: \SI{95}{\percent} clip-clustered bootstrap. Best in bold, second best
underlined (models only).}
\label{tab:human_chunks}
\small  %
\setlength{\tabcolsep}{1.5pt}
\begin{tabular}{lcc}
\toprule
Model, chunk score & ROC AUC [CI] & Catch @16\% \\
\midrule
SyncNet \cite{chung2016out}, L2 distance   & 0.664 [.581,.739] & 0.28 \\
SyncNet, LSE-C                              & 0.705 [.624,.778] & 0.38 \\
MTD-VocaLiST \cite{mtdvocalist2024}, sync prob. & 0.681 [.607,.754] & 0.30 \\
StableSyncNet \cite{li2024latentsync}, cosine & \underline{0.794} [.722,.859] & \underline{0.56} \\
\midrule
PVSync stage 3, window                      & 0.788 [.711,.859] & \underline{0.56} \\
\textbf{PVSync stage 3, worst phoneme}      & \textbf{0.836} [.781,.885] & \textbf{0.70} \\
\midrule
Single rater                                & --- & 0.59 \\
\bottomrule
\end{tabular}
\end{table}

\noindent\textbf{Generation-model level.} Thirteen talking-head video generation models produced
\num{324} clips, each shifted to its detected offset so that articulation alone is
judged. We asked \num{180} blind raters to score each clip from 1 to 5 on how
well the mouth matches the speech. A model's error rate is the
percentage of its phones scored below \num{0.28}, a threshold fixed on real
footage before the study. Ranked against the raters' mean scores,
the error rate agrees at Spearman \num{0.83}, where LSE-C
\cite{prajwal2020wav2lip} reaches \num{0.34}. By mean rating, five models score above 4 (good), six between 3 and 4 (mid) and two below 3 (bad). The
error rate ranks all five good models above every mid and bad model and
misorders only two of the twelve mid--bad
pairs (\cref{fig:space}b).

\noindent\textbf{Chunk level.} We sampled \num{400} chunks of \num{12} frames (\SI{480}{\ms}) from \num{240}
generated talking-head clips held out from all training. Ninety-six blind raters
saw each chunk with \num{13} frames of context on either side and
offset-corrected audio, and answered yes or no to whether the mouth matches the
speech. Each chunk got ten ratings on average, and their majority is its label,
\num{350} yes and \num{50} no. Compared with the rest of the panel's majority, a single rater catches
\SI{59}{\percent} of the chunks the panel flags and wrongly flags \SI{16}{\percent} of
the chunks it accepts. We fix every model's threshold at
this \SI{16}{\percent} false-flag rate and report the percentage of human-flagged
bad chunks it catches, along with its ROC AUC against the majority label
(\cref{tab:human_chunks}). We
compare our model with the baselines through two scores, the window score
of the whole chunk and the worst-phoneme score, the lowest per-phone score
(\cref{ssec:loc}) inside it. The window score reaches an ROC AUC of \num{0.788}
and catches \SI{56}{\percent} of the panel's flags. The worst-phoneme score
reaches \num{0.836} and catches \SI{70}{\percent}, above the single rater's
\SI{59}{\percent}. SyncNet and MTD-VocaLiST trail both of our scores,
StableSyncNet ties the window score, and no other model can score a chunk phone
by phone.

\section{Conclusion}
\label{sec:conclusion}

Clip-level lip-sync scores do not measure articulation. LSE-C ranks thirteen
talking-head video generation models at Spearman \num{0.34} against human raters,
where our per-phoneme error rate reaches \num{0.83}. PVSync scores lip-sync per
phoneme from one shared audio--visual embedding and calculates the offset to
within one frame with \SI{92.7}{\percent} accuracy, within \SI{0.4}{pp} of the
state of the art on LRS2 without training on it. Its viseme loss teaches the
embedding viseme awareness with no human labels, raising the ROC AUC for telling a
good phoneme--viseme match from a mismatch from \num{0.69} after stage~1 to
\num{0.96} after stage~2, \num{0.91} in the final model. On blind chunk ratings,
its worst-phoneme score matches a single rater's agreement with the panel.

\section{Acknowledgments}
Claude (Anthropic) was used under the authors' direction to edit and restructure
the text of all sections, to write and debug the evaluation code and the scripts
that render \cref{fig:space}, and to check the reference list; all content was
verified by the authors.

\section{Compliance with Ethical Standards}
The offset-recovery, viseme-mismatch and LRS2 experiments are numerical studies
on existing datasets for which no ethical approval was required. The rating
studies of \cref{ssec:human} involved paid crowdworkers recruited via Prolific
and administered through Mabyduck, who rated the perceptual lip-sync quality of
generated videos. Participants were adults (18+) who took part voluntarily after
receiving task instructions, and no ethics committee approval was required for
this minimal-risk perceptual evaluation task. Participants were compensated at \pounds18.55--\pounds19.67 per hour. The
authors received only the ratings, keyed by pseudonymous participant
identifiers. Other data comprised publicly available or proprietary research
datasets used under their respective licences, and videos generated by the
authors using commercial talking-head systems.

\bibliographystyle{IEEEbib}
\bibliography{refs}

\end{document}